\documentclass[runningheads]{llncs}
\usepackage[T1]{fontenc}
\usepackage{graphicx}
\usepackage{diagbox}
\usepackage{xcolor}
\usepackage{makeidx} 
\usepackage[colorlinks,linkcolor=blue]{hyperref}
\usepackage{amsmath}
\usepackage{booktabs}
\usepackage[misc]{ifsym} 
\usepackage{bbding}
\usepackage{amsfonts}
\usepackage{multirow}
\usepackage{xspace}
\usepackage{makecell}
\usepackage{enumitem} 
\usepackage{multirow}
\usepackage{tabularx}
\usepackage{array}
\usepackage{graphicx}
\usepackage{booktabs}
\usepackage{multirow}
\usepackage{subcaption}
\usepackage{caption}
\usepackage{graphicx,verbatim}
\begin{document}
\title{FrameONE: Hierarchical Motion Modeling for Universal Multi-View Echocardiographic Keyframe Detection}

\titlerunning{FrameONE}

\author{Rusi Chen\inst{1}\thanks{Rusi Chen, Yuhao Huang, and Hongyuan Zhang contributed equally.} 
\and Yuhao Huang\inst{1,2\star} 
\and Hongyuan Zhang\inst{2\star} 
\and Chao Tian\inst{3} 
\and Shunan Ji\inst{4,5} 
\and Yuhan Zhang\inst{1} 
\and Dong Ni\inst{1,3,4,5}\textsuperscript{(\Letter)}
} 

\institute{
\textsuperscript{$1$}Medical Ultrasound Image Computing (MUSIC) Lab,
Shenzhen University, Shenzhen, China\\
\email{nidong@szu.edu.cn} \\
\textsuperscript{$2$}Centre for Artificial Intelligence and Robotics (CAIR), Hong Kong Institute of Science \& Innovation, Chinese Academy of Sciences, Hongkong, China\\
\textsuperscript{$3$}School of Biomedical Engineering and Informatics, Nanjing Medical University, Nanjing, China\\
\textsuperscript{$4$}School of Artificial Intelligence, Shenzhen University, Shenzhen, China\\
\textsuperscript{$5$}National Engineering Laboratory for Big Data System Computing Technology, Shenzhen University, Shenzhen, China\\
}

\authorrunning{R.Chen et al.}
  
\maketitle              

\begin{abstract}
Accurate detection of end-systole (ES) and end-diastole (ED) frames is fundamental to echocardiographic assessment.
Existing methods are typically developed in a view-specific manner, depend on auxiliary annotations or intensive visual modeling, which limits their generalizability.
In multi-view modeling, keyframe detection is driven by shared cardiac motion, yet large appearance differences and motion patterns make unified modeling challenging.
To address these issues, we propose \textbf{FrameONE}, a unified end-to-end framework for multi-view echocardiographic keyframe detection.
FrameONE introduces a \textbf{Hierarchical Motion Modeling} strategy: 
an intra-view multi-task learning reduces appearance bias and promotes motion-focused representations within each view; an inter-view general motion learning module further separates view-agnostic dynamics from view-specific patterns, enabling shared yet flexible motion representation learning across views.
Extensive experiments on 25,872 videos spanning four standard views demonstrate that FrameONE achieves state-of-the-art keyframe detection accuracy with strong cross-view generalization.
Code is available at \url{https://github.com/szuboy/FrameONE}.


\end{abstract}

\section{Introduction}
\begin{figure}[!t]
	\begin{center}
    \includegraphics[width=1\columnwidth]{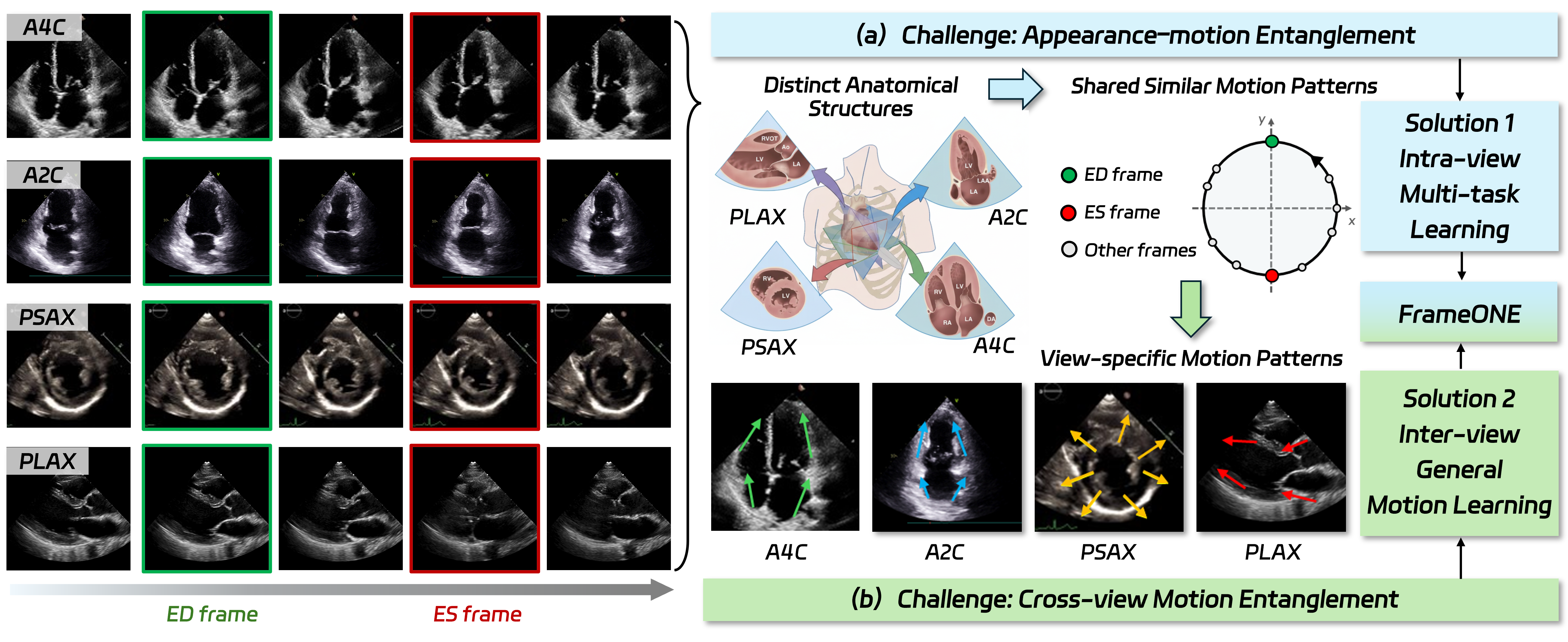}
	\end{center}
    \caption{
    Overview of the two core challenges in multi-view cardiac keyframe detection and the corresponding solutions in FrameONE. Four echocardiographic views are: apical four-chamber (A4C), apical two-chamber (A2C), parasternal long-axis (PLAX), and parasternal short-axis (PSAX).}
	\label{fig:task}
\end{figure}

The accurate identification of end-systole (ES) and end-diastole (ED) frames is fundamental to echocardiographic assessment, as clinical parameters are typically derived from these keyframes across multiple standardized views~\cite{lang2015recommendations}.  
However, the anatomical features defining these keyframes vary significantly among different views, hindering consistent and reliable identification (Fig.~\ref{fig:task}a)~\cite{mcleod2015spatio}. 
Besides, manual assessment is labor-intensive and susceptible to intra- and inter-observer variability.
Therefore, these challenges motivate the development of view-agnostic automated keyframe detection model.

Existing deep learning approaches have shown satisfactory progress in intelligent heart ultrasound analysis~\cite{yu2024explainable,zhou2024heartbeat,chen2025mtcnet,zhou2026onuvs,zhou2026ctrl}. 
However, for ES/ED frame detection, most methods are predominantly designed for single-view settings and often rely on view-specific assumptions~\cite{lane2021multibeat,farhad2023cardiac,tasken2023automated,chen2025semi,li2023echoefnet}. 
Several supervised methods incorporate additional anatomical or clinical annotations.
For instance, 
Feng et al.~\cite{feng2024bayesian} jointly detect keyframes and landmarks, but rely on view-specific landmark motion that may limit cross-view generalization.
Similarly, Reynaud et al.~\cite{reynaud2021ultrasound} adopt a multi-task framework that requires extra ejection fraction supervision, increasing annotation demands.
Yang et al.~\cite{yang2025latent} introduced an unsupervised decoupling method, but it is strictly tailored to the A4C view by assuming orthogonal motion directions specific to A4C anatomy.
Furthermore, Lu et al.~\cite{lu2025optical} integrate optical flow to explicitly model cardiac dynamics, at the expense of increased computational overhead and sensitivity to redundant visual information.
Overall, previous methods remain largely view-dependent and do not explicitly leverage shared physiological patterns or complementary information across multiple views, limiting their clinical applicability.

In this work, instead of view-specific designs, we propose an end-to-end unified framework for detecting ES and ED frames across multiple echocardiographic views, named \textbf{FrameONE}.
As shown in Fig.~\ref{fig:task}, developing such a unified model faces two fundamental challenges.
1) Although different views share the same cardiac rhythm, they exhibit distinct anatomical structures. 
Consequently, models may tend to focus on view-specific appearance features rather than motion dynamics.
2) Even under the same cardiac cycle, motion patterns differ across views, such as longitudinal contraction in A4C and radial contraction in PSAX.
Overall, the main contributions of this work are as follows:
\begin{itemize}[label=\textbullet, itemsep=0.3em] 
\item \textbf{A pioneering unified framework}. We present FrameONE, to the best of our knowledge, is the first attempt to explore multi-view echocardiographic keyframe detection without view-specific design.

\item We introduce \textbf{Hierarchical Motion Modeling (HMM)}, including Intra-view Multi-task Learning (IML) and an Inter-view General Motion Learning (IGM) strategies. IML disentangles the representation of motion and structural features, IGM decouples motion into view-agnostic and view-specific patterns, enabling shared motion representation learning across views.

\item Extensive experiments across four echocardiographic views demonstrate that FrameONE outperforms existing methods on most evaluation metrics.

\end{itemize}

\section{Method}

As illustrated in Fig.~\ref{fig:framework}, given an echocardiographic video $\mathbf{X}=\{\mathbf{x}_1,\dots,\mathbf{x}_T\}$ from arbitrary view (A4C/A2C/PLAX/PSAX), FrameONE detects ES and ED frames by regressing a continuous score for each frame, where 0 corresponds to ES and 1 to ED.
The framework employs a shared encoder to extract inter-view motion features, followed by an IGM module that disentangles view-agnostic and view-specific representations.
These features are then fed into a dual-decoder for ES/ED phase regression and view classification.

\subsection{Intra-view Multi-task Learning}
Different echocardiographic views have distinct appearances, which can make the model rely on visual cues rather than cardiac motion.
To address this issue, we introduce IML to jointly optimize keyframe regression and view classification by a dual-decoder learning paradigm. 
This design encourages the network to focus on motion patterns that are consistent for keyframe detection.

\begin{figure}[!t]
	\begin{center}
    \includegraphics[width=1.0\textwidth]{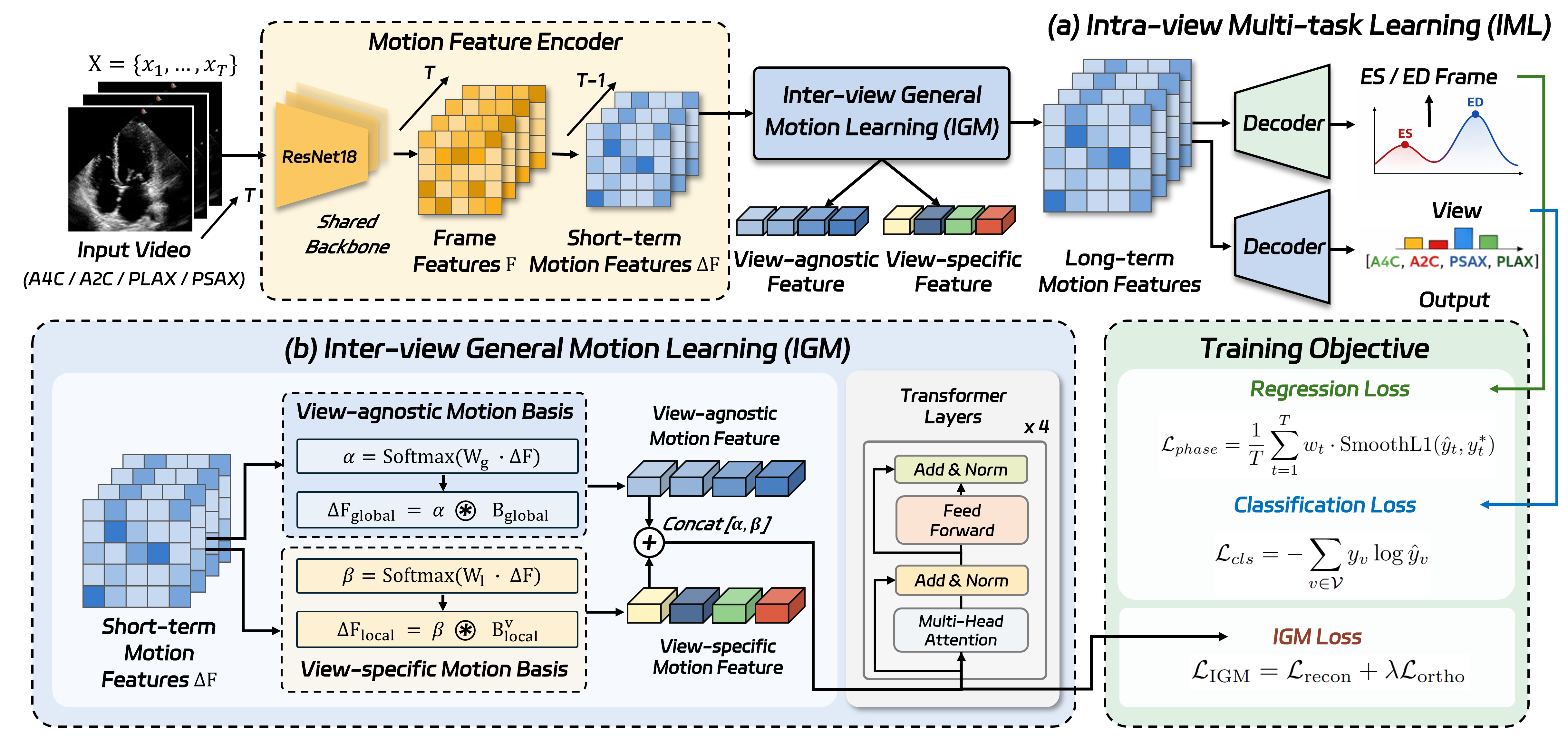}
	\end{center}
    \caption{Overview of proposed FrameONE. }
	\label{fig:framework}
\end{figure}

As illustrated in Fig.~\ref{fig:framework}a, each frame $\mathbf{x}_t$ is encoded by a shared ResNet18~\cite{he2016deep} backbone with a fully connected layer 
$\mathbf{f}_t = \mathcal{E}(\mathbf{x}_t) \in \mathbb{R}^{256}$.
However, static frame features alone cannot capture the dynamic contraction-relaxation process essential for identifying ES/ED.
While optical flow for model motion may introduce computational overhead and is sensitive to speckle noise and probe motion.
Here, we extract short-term motion representations by applying a learnable 1D temporal convolution over the feature sequence $\mathbf{F} = [\mathbf{f}_1, \dots, \mathbf{f}_T]^\top \in \mathbb{R}^{T \times d}$:
\begin{equation}
\Delta \mathbf{f}_t = \text{Conv1D}(\mathbf{F}_t), \quad t = 1, \dots, T-1,
\end{equation}
where the convolution operates along the temporal axis with kernel size $k$.
This formulation aligns naturally with the task: ES and ED correspond to motion extrema (sign changes in $\Delta \mathbf{f}_t$) rather than appearance extrema.

The multi-task decoder consists of two heads: (i) a regression head that outputs frame-level scores $\hat{y}_t \in [0,1]$, and (ii) a view classification head that predicts the imaging view.
The regression loss is a weighted Smooth L1 loss:
\begin{equation}
\mathcal{L}_{phase} = \frac{1}{T} \sum_{t=1}^{T} w_t \cdot \text{SmoothL1}(\hat{y}_t, y_t^*),
\end{equation}
where the per-frame weights $w_t$ assign higher importance to frames near the ES/ED boundaries.
To obtain $w_t$, we construct a Gaussian keyframe-proximity map $w_{\text{gauss}}(t)$ by centering
1D Gaussian kernels with $\sigma = 2$ at the ES and ED frame indices and aggregating their responses
over time. The final weighting is defined as $w_t = 1 + w_{\text{gauss}}(t)$, which up-weights the
keyframe neighborhoods and encourages the model to prioritize regression accuracy around ES/ED.
The view classification loss is defined as the standard cross-entropy:
\begin{equation}
\mathcal{L}_{cls} = -\sum_{v \in \mathcal{V}} y_v \log \hat{y}_v,
\end{equation}
where $\mathcal{V}$ denotes the set of view categories, $y_v \in \{0,1\}$ is the one-hot ground-truth
indicator for class $v$, and $\hat{y}_v \in [0,1]$ is the predicted probability for view $v$
produced by the classification head.

\subsection{Inter-view General Motion Learning}
Cardiac views share a common contraction-relaxation rhythm that reflects global cardiac physiology. 
However, each view also presents its own motion characteristics (see Fig.~\ref{fig:task}b).  
Yet, learning motion features directly from multi-view data often mix view-specific motion with the global contraction–relaxation pattern, which obscures the underlying signal and leads to suboptimal results.

To address this issue, we disentangle motion into a shared rhythm component and a view-dependent residual component (see Fig.~\ref{fig:framework}b).
The core idea is to factorize representations into two spaces: a view-agnostic global subspace and a view-specific local subspace.
Specifically, let $\mathbf{B}_{global}$ be a global basis matrix whose $K_g$ columns capture universal cardiac dynamics shared across all views, and $\mathbf{B}_{local}^{v}$ be a view-specific basis matrix whose $K_l$ columns model local motion variations for view $v$.
The reconstruction motion is then decomposed as:
\begin{equation}
\Delta \hat{\mathbf{f}}_t
= \Delta \hat{\mathbf{f}}_t^{g} + \Delta \hat{\mathbf{f}}_t^{l}
= \boldsymbol{\alpha} \cdot \mathbf{B}_{global}
+ \boldsymbol{\beta} \cdot \mathbf{B}_{local}^{v},
\end{equation}
where $\boldsymbol{\alpha}$ and $\boldsymbol{\beta}$ represent projection coefficients in the global and local subspaces. 
We then apply a softmax function to encourage sparse basis activation:
\begin{equation}
\boldsymbol{\alpha} = \mathrm{softmax}(\mathbf{W}_g^\top \Delta \mathbf{f}_t), 
\quad
\boldsymbol{\beta} = \mathrm{softmax}(\mathbf{W}_l^\top \Delta \mathbf{f}_t),
\end{equation}
where $\mathbf{W}_g \in \mathbb{R}^{d \times K_g}$ and $\mathbf{W}_l \in \mathbb{R}^{d \times K_l}$ are learnable projection matrices. 
This formulation encourages the model to separate shared physiological dynamics from view-dependent motion patterns. 
Then, the global and local coefficient vectors are concatenated as 
$\mathbf{z}_t = \text{Concat}(\boldsymbol{\alpha}; \boldsymbol{\beta})$, 
forming a compact motion descriptor at each time step. 
Over a cardiac cycle, the sequence $\mathbf{z}_t$ traces a low-dimensional trajectory that characterizes contraction and relaxation dynamics. 
Since ES and ED correspond to extrema of this cyclic process rather than local motion magnitude, we model the global temporal structure of this trajectory using a 4-layer transformer temporal encoder, yielding contextualized representations.
Then, they are fed to the multi-task decoder to regress scores and classify view types.

Finally, in order to ensure proper disentanglement between shared and view-specific dynamics, we adopt a reconstruction objective $\mathcal{L}_{\mathrm{recon}}$ together with a cross-subspace orthogonality constraint $\mathcal{L}_{\mathrm{ortho}}$ to formulate the loss function as:

\begin{equation}
\begin{aligned}
\mathcal{L}_{\mathrm{IGM}}
&=  \mathcal{L}_{\mathrm{recon}}
 + \lambda\mathcal{L}_{\mathrm{ortho}} \\
&= \frac{1}{T-1} \sum_{t=1}^{T-1}
\|\Delta \hat{\mathbf{f}}_t - \Delta \mathbf{f}_t\|_2^2
+ \lambda \|\mathbf{U}_g^\top \mathbf{U}_l^{v}\|_F^2
+ \lambda \|\mathbf{z}_t\|_1,
\end{aligned}
\end{equation}
where $\lambda$ controls the orthogonality constraints ($\lambda=0.1$ based on empirical observations).
$\mathcal{L}_{\mathrm{IGM}}$ mitigates leakage between global rhythm and view-dependent components, which improves identifiability and stabilizes training.

\subsection{Training Objective and Inference}
\subsubsection{Training.} The overall FrameONE is trained in an end-to-end manner by jointly optimizing the phase regression objective and multiple auxiliary regularization terms.
The overall loss function is defined as:

\begin{equation}
\mathcal{L} 
= \mathcal{L}_{phase} 
+ \lambda_{cls} \mathcal{L}_{cls} 
+ \lambda_{\text{IGM}} \mathcal{L}_{\text{IGM}},
\end{equation}
where $\lambda_{cls} = \lambda_{\text{IGM}}= 0.1$. The supervision signal is provided by $\mathcal{L}_{phase}$,
while the auxiliary view classification loss 
$\mathcal{L}_{cls}$ guides the shared encoder to preserve 
view-discriminative anatomical information.
In addition,
the IGM regularization term further ensures more stable optimization.

\subsubsection{Inference.} 
During testing, we use a sliding-window strategy with 50\% overlap to process videos of arbitrary length and average predictions in overlapping regions. 
The score curve is thresholded at 0.5 to obtain candidate segments. 
Consecutive candidates are grouped, and the peak frame within each group is selected as the local extremum using a 5-frame window. 
If no candidate is detected, the global minimum and maximum of the score curve are used as fallback predictions.

\section{Experiments and Results}

\begin{table}[!t]
\renewcommand{\arraystretch}{1.05}
\centering
\caption{
Comparison results with other state-of-the-art keyframe detection methods across four views.
$\text{FrameOne} {\scriptsize\text{(Single)}}$ means model trained with single view data.
The best results are colored in \textcolor{blue}{blue}, and the second are \underline{underlined}.
}
\setlength{\tabcolsep}{7pt}
\label{tab:result_table}
\footnotesize
\resizebox{\textwidth}{!}{
\begin{tabular}{l|l|lll|ll}
\hline
\multirow{2}{*}{View}
& \multirow{2}{*}{Method}
& \multicolumn{3}{c|}{Single-cycle}
& \multicolumn{2}{c}{Multi-cycle}
\\ \cline{3-7}
&                         & MAE $\downarrow$ & RMSE$\downarrow$ & R$^2$$\uparrow$ & FP$\downarrow$ &  FN$\downarrow$ \\ \hline
\multirow{7}{*}{A4C}  & DeepPhase~\cite{farhad2023cardiac}{\scriptsize(SR, 2023)}           & 7.707  & 24.581 & 0.619 & 9.556 & 1.737 \\
                      & CVC~\cite{tasken2023automated}{\scriptsize(AIM, 2023)}              & 7.606  & 18.566 & 0.783 & 3.917 & 1.453 \\
                      & ResLSTM~\cite{lane2021multibeat}{\scriptsize(CBM, 2021)}            & 9.844  & 12.115 & 0.907 & 5.613 & 2.762 \\
                      & UVT~\cite{reynaud2021ultrasound}{\scriptsize(MICCAI, 2021)}         & 2.599  & 17.450 & 0.867 & 5.242 & 3.144 \\
                      & OFM~\cite{lu2025optical}{\scriptsize(TMI, 2025)}                    & \underline{2.832}  & \underline{3.527}  & \underline{0.994} & 3.762 & 1.149 \\
                      & FrameONE {\scriptsize(Single)}                                       & 2.883  & 3.980  & 0.993 & \underline{1.925} & \textcolor{blue}{0.803} \\
                      & \textbf{FrameONE {\scriptsize(Ours)}}  & \textcolor{blue}{2.339}  & \textcolor{blue}{2.974}  & \textcolor{blue}{0.995} & \textcolor{blue}{1.920} & \textcolor{blue}{0.803} \\ \hline
\multirow{7}{*}{A2C}  & DeepPhase~\cite{farhad2023cardiac}{\scriptsize(SR, 2023)}           & 14.679 & 28.569 & 0.286 & 9.952 & 1.809 \\
                      & CVC~\cite{tasken2023automated}{\scriptsize(AIM, 2023)}              & 14.771 & 27.360 & 0.342 & 3.904 & 1.785 \\
                      & ResLSTM~\cite{lane2021multibeat}{\scriptsize(CBM, 2021)}            & 3.534  & 4.340  & 0.985 & 7.904 & 1.666 \\
                      & UVT~\cite{reynaud2021ultrasound}{\scriptsize(MICCAI, 2021)}         & 6.302  & 12.817 & 0.893 & 5.404 & 2.904 \\
                      & OFM~\cite{lu2025optical}{\scriptsize(TMI, 2025)}                    & \textcolor{blue}{3.077}  & \textcolor{blue}{3.972}  & \textcolor{blue}{0.989} & \underline{2.738} & \textcolor{blue}{1.428} \\
                      & FrameONE {\scriptsize(Single)}       & 3.464  & 6.339  & 0.972 & \textcolor{blue}{2.450} & \underline{1.800} \\
                      & \textbf{FrameONE {\scriptsize(Ours)}} & \underline{3.121}  & \underline{4.854}  & \underline{0.984} & 2.974 & 2.325 \\ \hline
\multirow{7}{*}{PSAX} & DeepPhase~\cite{farhad2023cardiac}{\scriptsize(SR, 2023)}           & 6.204  & 17.589 & 0.782 & 7.731 & 1.052 \\
                      & CVC~\cite{tasken2023automated}{\scriptsize(AIM, 2023)}              & 6.347  & 13.202 & 0.871 & 2.731 & 1.097 \\
                      & ResLSTM~\cite{lane2021multibeat}{\scriptsize(CBM, 2021)}            & 7.887  & 10.664 & 0.920 & 3.253 & 1.276 \\
                      & UVT~\cite{reynaud2021ultrasound}{\scriptsize(MICCAI, 2021)}         & 6.032  & 15.193 & 0.827 & 3.388 & 1.582 \\
                      & OFM~\cite{lu2025optical}{\scriptsize(TMI, 2025)}                    & 2.970  & \textcolor{blue}{5.136}  & \underline{0.978} & \textcolor{blue}{2.395} & \textcolor{blue}{0.850} \\
                      & FrameONE {\scriptsize(Single)}                                       & \underline{2.891}  & \underline{5.274}  & \textcolor{blue}{0.979} & 2.719 & 1.136 \\
                      & \textbf{FrameONE {\scriptsize(Ours)}} & \textcolor{blue}{2.520}  & 6.171  & 0.971 & \underline{2.424} & \underline{1.113} \\ \hline
\multirow{7}{*}{PLAX} & DeepPhase~\cite{farhad2023cardiac}{\scriptsize(SR, 2023)}           & 7.204  & 20.828 & 0.678 & 9.740 & 2.030 \\
                      & CVC~\cite{tasken2023automated}{\scriptsize(AIM, 2023)}              & 6.656  & 13.947 & 0.861 & 5.010 & 1.955 \\
                      & ResLSTM~\cite{lane2021multibeat}{\scriptsize(CBM, 2021)}            & 6.722  & 8.934  & 0.943 & 7.450 & 2.625 \\
                      & UVT~\cite{reynaud2021ultrasound}{\scriptsize(MICCAI, 2021)}         & 2.992  & 17.339 & 0.932 & 5.090 & 3.185 \\
                      & OFM~\cite{lu2025optical}{\scriptsize(TMI, 2025)}                    & 3.071  & 5.026  & 0.993 & 3.920 & 1.725 \\
                      & FrameONE {\scriptsize(Single)}                                       & \underline{2.509}  & \textcolor{blue}{4.150}  & \textcolor{blue}{0.996} & \textcolor{blue}{2.573} & \textcolor{blue}{1.615} \\
                      & \textbf{FrameONE {\scriptsize(Ours)}} & \textcolor{blue}{2.250}  & \underline{4.376}  & \underline{0.995} & \underline{2.647} & \underline{1.647} \\ \midrule
\multirow{2}{*}{Avg}
& OFM~\cite{lu2025optical}{\scriptsize(TMI, 2025)}
& 2.907 & 4.151 & \textcolor{blue}{0.990} & 3.438 & 1.173 \\
& \textbf{FrameONE {\scriptsize(Ours)}}
& \textcolor{blue}{2.386} & \textcolor{blue}{3.994} & 0.989
& \textcolor{blue}{2.176} & \textcolor{blue}{1.042} \\
\hline
\end{tabular}
}
\end{table}

\vspace{-8pt}
\subsubsection{Implementation Details.}
Our model was implemented with PyTorch on two NVIDIA GeForce RTX 4090 GPUs. 
All input frames were resampled to \(128 \times 128\). 
We followed~\cite{reynaud2021ultrasound} to duplicate the transition frames between the two labeled frames and append them after the last annotation.
During training, we employed a random start fixed-length sampling strategy to extract 50 frames from each video.
AdamW optimizer~\cite{loshchilov2019decoupledweightdecayregularization} was used with an initial learning rate of 5e-3 and weight decay of 1e-5.
We used a cosine annealing learning rate scheduler with 5 warmup epochs.
The batch size was 32, and the total training epoch was 100.

\begin{figure}[!t]
	\begin{center}
    \includegraphics[width=1\textwidth]{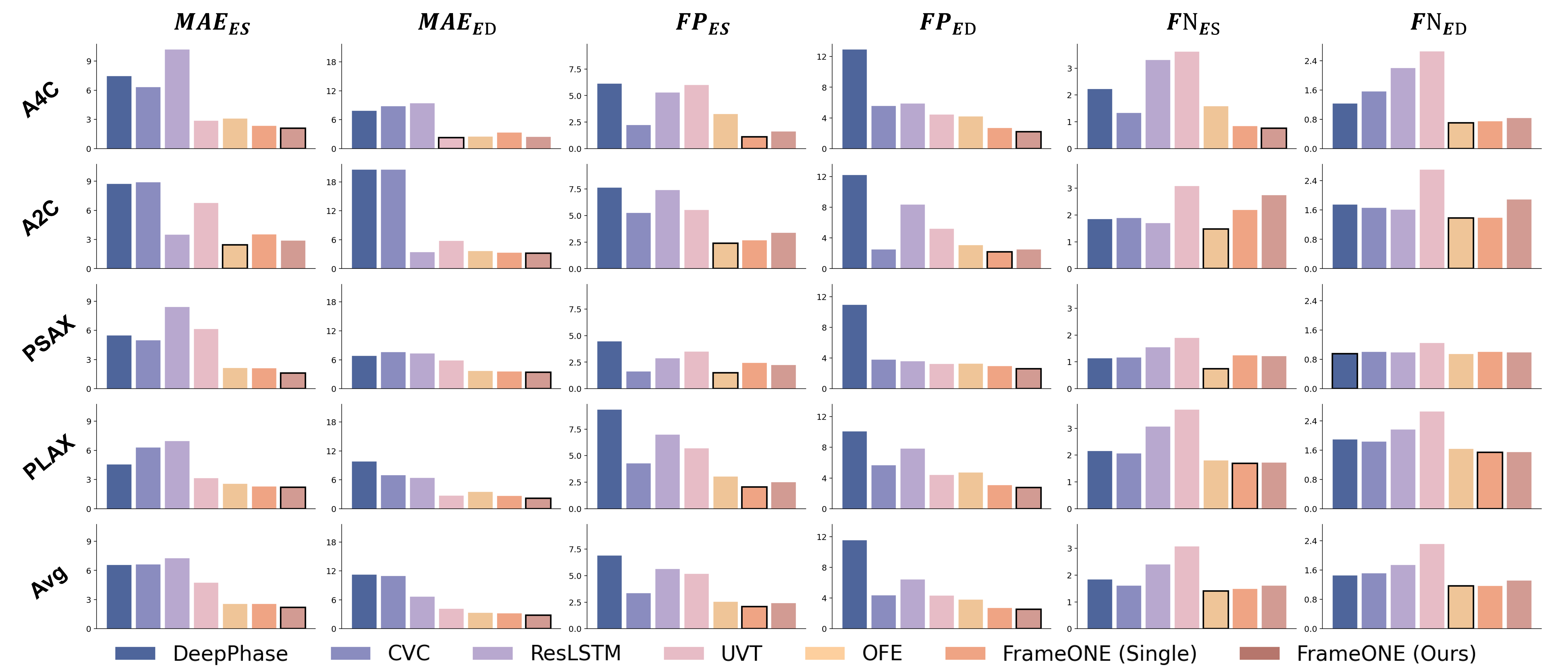}
	\end{center}
    \caption{Quantitative comparison of ES/ED detection across four views.}
	\label{fig:visual_result}
\end{figure}

\subsubsection{Datasets and Evaluation Metrics.}  
We evaluate our method on a large cross-population echocardiographic cohort covering four cardiac views. The public part includes A4C from EchoNet-Dynamic~\cite{ouyang2019echonet}, PLAX from EchoNet-LVH~\cite{duffy2022high}, and PSAX from Echo-pediatric~\cite{reddy2023video}, while the A2C view is from our private dataset. 
In total, we collected 25,872 samples, comprising 9,926 A4C, 11,021 PLAX, 4,475 PSAX, and 450 A2C videos. 
The overall videos are split into 20,338, 3,367, and 2,167 for training, validation, and testing, respectively.
Since public datasets provide only one ES/ED pair label per video.
To evaluated method performance in read-world clinical settings, we additionally annotated 50–100 videos per view with ES/ED labels across 3-5 cardiac cycles by experienced sonographers. 
Thus, we evaluate the model under \textit{single-cycle annotation} and \textit{multi-cycle annotation} settings.
 The former uses MAE, RMSE, and $R^2$ as metrics, while the latter assesses FP and FN rates, defined as $N_{\mathrm{FP}} / N_{\mathrm{pred}}$ and $N_{\mathrm{FN}} / N_{\mathrm{GT}}$, respectively, where $N_{\mathrm{FP}}$ and $N_{\mathrm{FN}}$ denote unmatched predictions and missed ground truths. 
 All metrics are reported at the frame level.

\subsubsection{Comparison Study.}
As shown in Table~\ref{tab:result_table}, FrameONE consistently achieves the best performance across all four echocardiographic views.
Compared with the state-of-the-art OFM, FrameONE further reduces the overall MAE from 2.907 to 2.386 frames (p<0.001) and decreases FP from 3.438 to 2.176 (p<0.001).
In addition, FrameONE achieves a mean MAE of 2.339 frames on A4C, yielding 17.4\% and 10.0\% lower errors than OFM and UVT, respectively.
Similar improvements are observed on the challenging PSAX and PLAX with less distinctive anatomies.
In Fig.~\ref{fig:visual_result}, we further report detailed ES and ED metrics, demonstrating that FrameONE achieves the best performance in most cases.
For the competing methods, DeepPhase and CVC underperform due to the missing temporal contexts of CNNs.
UVT lacks explicit motion decomposition and depends on auxiliary ejection fraction supervision.
ResLSTM and OFM rely on complex temporal modeling (e.g., optical flow), incurring high computational cost.
Moreover, FrameONE achieves 246.6 frame per second (FPS) with only 14.35M parameters, outperforming OFM (132.1 FPS, 43.93M) and ResLSTM (49.2 FPS, 29.81M).
This further confirms the effectiveness of our method.

The multi-label testing set supports robustness evaluation, where FrameONE attains substantially lower FP (1.920 \textit{vs.} 3.762/5.242) and FN (0.803 \textit{vs.} 1.149/3.144) than the strong OFM and UVT (see Table~\ref{tab:result_table}).
It is noted that FrameONE (Single) trained independently per view already outperforms all baselines, while our unified multi-view model further reduces MAE by $\sim$0.3-0.5 across views, demonstrating the necessity of our designed joint learning.

\begin{table}[!t]
  \centering
  \setlength{\tabcolsep}{4pt}

    \begin{minipage}[t]{0.47\textwidth}
        \centering
        \footnotesize
        \renewcommand{\arraystretch}{1.1}
        \caption{Ablation study results.}
        \label{tab:ablation}
        \begin{tabular}{c c c c}
            \toprule
            \multirow{2}{*}{IML} 
            & \multicolumn{2}{c}{IGM} 
            & \multirow{2}{*}{MAE$\downarrow$} \\
            \cmidrule(lr){2-3}
            & Global Basis & Local Basis & \\
            \midrule
            $\times$      & $\times$ & $\times$ & 3.093 \\
            $\checkmark$  & $\times$ & $\times$ & 2.700 \\
            $\checkmark$  & $\checkmark$ & $\times$ & 2.470 \\
            $\checkmark$  & $\times$ & $\checkmark$ & 2.480 \\
            $\checkmark$  & $\checkmark$ & $\checkmark$ & \textbf{2.362} \\
            \bottomrule
        \end{tabular}
    \end{minipage}
  \hfill
  \begin{minipage}[t]{0.45\textwidth}
    \centering
    \vspace{0pt}
    \includegraphics[width=\linewidth]{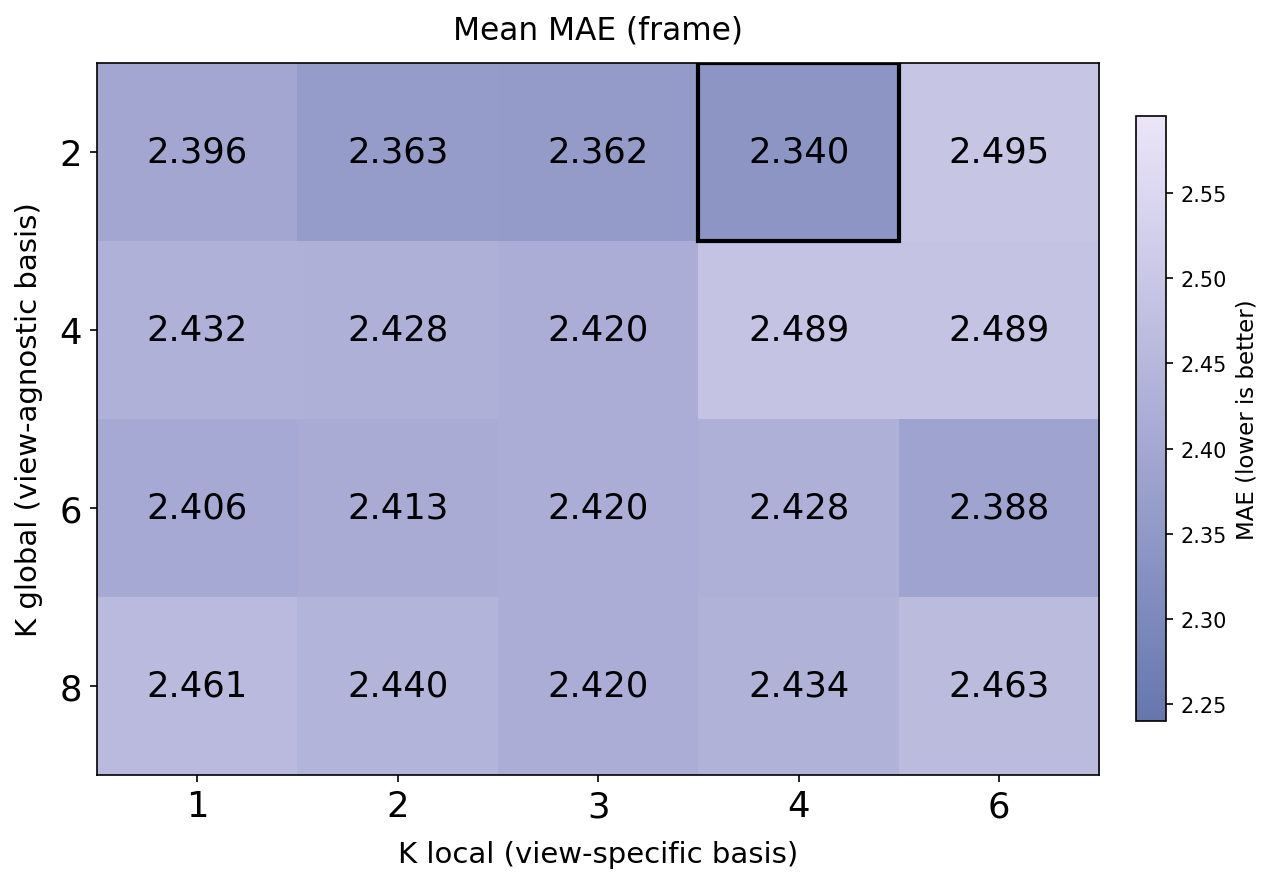}
    \captionof{figure}{Effect of basis number $K_{\text{g}}$ and $K_{l}$ in the IGM module.}
    \label{fig:ablation_fig}
  \end{minipage}

\end{table}

\subsubsection{Ablation Study.} \label{sec:ablation_study}
Table~\ref{tab:ablation} comprehensively tests each component of HMM, starting from a baseline model that directly predicts keyframes from raw frame features.
First, adding IML reduces mean MAE from 3.093 to 2.700 frames, validating the importance of explicit inter-view learning.
Incorporating the global basis further reduces MAE to 2.470 frames, demonstrating that capturing universal cardiac rhythm patterns across views benefits keyframe detection.
Besides, the local basis yields a 0.613-frame reduction in MAE, suggesting the importance of view-dependent motion modeling. 
With all components, FrameONE achieves the best performance, validating the synergistic benefits of our hierarchical design.
In Fig.~\ref{fig:ablation_fig}, we carefully analyze the model sensitivity to the basis number.
The mean MAE varies only slightly (2.340–2.495) across different settings, indicating that extensive parameter tuning is not required for FrameOne.

\section{Conclusion}
In this work, we presented FrameONE, 
the first unified framework for multi-view echocardiographic keyframe 
detection without view-specific design.
The key idea of proposed method is hierarchical motion modeling: 
IML disentangles motion representations from view-dependent appearance 
within each view, while IGM further factorizes the learned motion 
into shared cardiac rhythms and view-specific subspaces, 
forming a tightly coupled hierarchy.
Extensive experiments across four echocardiographic views 
demonstrate state-of-the-art performance.
Future work will explore extending FrameONE to additional cardiac views 
and integrating it with downstream functional assessment tasks.

\begin{credits}
\subsubsection{\ackname} This work was supported by the grant from National Natural Science Foundation of China (12326619, 62572324);  
Frontier Technology Development Program of Jiangsu Province (BF2024078);
and Science and Technology Planning Project of Guangdong Province (2023A0505020002).
\subsubsection{\discintname}
The authors have no competing interests to declare that are relevant to the content of this article.
\end{credits}

\bibliographystyle{splncs04}
\bibliography{paper-3530}

\end{document}